# Can Webcam Gaze Constrain Mesa-Objectives in Driving Models? An Instrument Precision Analysis

Lennox Anderson[1,2], Ahmed Boutar[1], Jonah Mulcrone[2], Tal Erez[1]

[1]Duke University, Durham, NC [2]ONYX AI LLC

Correspondence: Lennox Anderson (lennoxanderson@onyxaillc.com)

**Abstract**

Current hazard detection systems in autonomous driving may develop mesa objectives, learned internal goals that achieve high training performance through spurious correlations rather than genuine hazard recognition. We investigate whether human gaze patterns, captured via webcam-based eye tracking (WebGazer.js), can serve as privileged information to constrain mesa-objective formation. We collected 137,663 frame-level gaze samples synchronized with hazard annotations across 388 real dashcam clips, then test this hypothesis across two calibration protocols (9-point/45-click and 11-point/440-click), two model architectures (Random Forest and causal Transformer), and five random seeds per experiment with paired $t$-tests. No experiment yields a statistically significant improvement from gaze ($p = 0.919$, $0.578$, and $0.667$ respectively). A geometric analysis reveals the root cause: WebGazer's reported error (∼130–257 px depending on configuration) exceeds 93% of detected hazard object sizes (median 36 px), rendering object-level gaze attribution physically impossible at this instrument precision.

## 1. Introduction

Autonomous driving systems must identify and respond to hazardous situations, but current approaches can develop *mesa objectives*, internal goals learned during training that differ from the intended base objective (Hubinger et al., 2019). A model might learn "flag all moving objects" rather than genuinely understanding hazard assessment.

We hypothesize that incorporating human gaze patterns during training provides *process-level supervision* that constrains the space of learnable mesa objectives. This framing connects to the *Learning Using Privileged Information* (LUPI) paradigm (Vapnik & Vashist, 2009): gaze is available during training but not at deployment, analogous to a teacher providing hints that the student must eventually internalize (Hinton et al., 2015). Where humans look reveals the causal mechanisms they use for hazard assessment, potentially forcing models to learn genuine risk features rather than proxy correlations.

To test this, we built an end-to-end data collection and modeling pipeline: dashcam footage from a Tesla Model 3, a webcam-based eye tracking simulation platform, a four-stage ETL pipeline, and two model architectures. Our investigation systematically eliminates three potential bottlenecks (calibration quality, model complexity, and the instrument itself), arriving at a definitive conclusion about why webcam gaze cannot currently improve hazard detection.

This research evolved through three phases: from mesa-objective alignment, through knowledge distillation, to a data quality investigation. Each pivot was driven by the evidence. Rather than forcing a positive conclusion from a single promising seed, we followed the data to its root cause.

## 2. Related Work

**Gaze as privileged information.** The LUPI paradigm (Vapnik & Vashist, 2009) formalizes the use of additional information available only during training, with Lopez-Paz et al. (2016) proving that knowledge distillation (Hinton et al., 2015) and privileged information are instances of the same framework. Gaze has been successfully used as privileged information in several domains: Karessli et al. (2017) used gaze embeddings for zero-shot image classification at CVPR, and Saran et al. (2021) showed that a gaze-based auxiliary loss improved imitation learning by 95–390% across 20 Atari games without adding learnable parameters. These successes, however, relied on research-grade eye trackers with sub-degree accuracy.

**Driver attention datasets.** Several benchmarks have demonstrated that high-quality gaze data can predict driver attention and anticipate hazards. Alletto et al. (2016) introduced DR(eye)VE (74 videos, SMI glasses at high precision), later extended by Palazzi et al. (2018) with a multi-branch deep architecture. Xia et al. (2018) built BDD-A (1,232 braking event videos) using an EyeLink 1000 at 1000 Hz. Fang et al. (2019) created DADA-2000 (2,000 clips, 54 accident types) with an SMI RED250 at 250 Hz, directly asking whether accidents can be predicted from driver attention. All three datasets use dedicated infrared hardware with sub-degree accuracy, a critical distinction from our webcam-based approach.

**Webcam eye tracking limitations.** WebGazer.js (Papoutsaki et al., 2016) enables browser-based gaze tracking without specialized hardware, achieving mean errors of 130–257 px. Semmelmann & Weigelt (2018) measured webcam accuracy at ∼4° visual angle in controlled lab settings, degrading further online.

Krafka et al. (2016) trained a CNN (iTracker) on 2.5M frames from 1,450 participants, achieving 1.7 cm error on phones, demonstrating that even with massive training data, consumer-device precision remains far below infrared. Wisiecka et al. (2022) confirmed that calibration quality directly determines downstream data usability.

**Hazard perception.** Horswill (2016) established that hazard perception is a learnable skill predictive of crash risk. Underwood et al. (2003) showed experienced drivers display wider horizontal scanning than novices, and Crundall et al. (2012) found that these differences manifest in subtle fixation placement variations (fractions of a degree) that require high-precision eye tracking to resolve. This establishes both that the gaze signal during hazards is information-rich and that it requires sub-degree spatial resolution to capture.

**Data quality and noisy labels.** Northcutt et al. (2021) found that even 3.4% average label error across major benchmarks destabilizes model rankings. Budach et al. (2022) systematically showed that noise, incompleteness, and inconsistency all degrade ML performance. Instrument-precision-induced noise (as with webcam gaze) creates spatially correlated errors rather than random noise, which is qualitatively more harmful to learning.

**Mesa-objectives.** Hubinger et al. (2019) formalized how gradient descent can produce models with internal objectives that differ from the training objective. Langosco et al. (2022) provided the first empirical demonstrations of goal misgeneralization in deep RL, showing agents retain capabilities out-of-distribution but pursue wrong goals. When privileged information is noisy, models may learn proxy objectives (e.g., “attend to large objects”) rather than the intended one (“attend where drivers look during hazards”).

# 3. Data Collection Pipeline

## 3.1. From Static Images to Video

An earlier version of this platform used static images from the Cityscapes dataset, collecting gaze data as participants identified hazards in individual frames. Figure 1 shows the resulting saliency maps across three analysis modes: aggregate attention (all viewers combined), viewer-separated attention (each color represents a different participant’s gaze cluster), and temporal attention (green indicates early fixations, red indicates late fixations within each viewing session). While these maps demonstrate that webcam gaze can produce spatially meaningful attention patterns on static stimuli, they cannot capture the sequential gaze dynamics that characterize real-world hazard detection: how attention shifts over time as a hazard develops. This limitation motivated the transition to continuous dashcam video, where temporal gaze trajectories during hazard onset become the primary signal of interest.

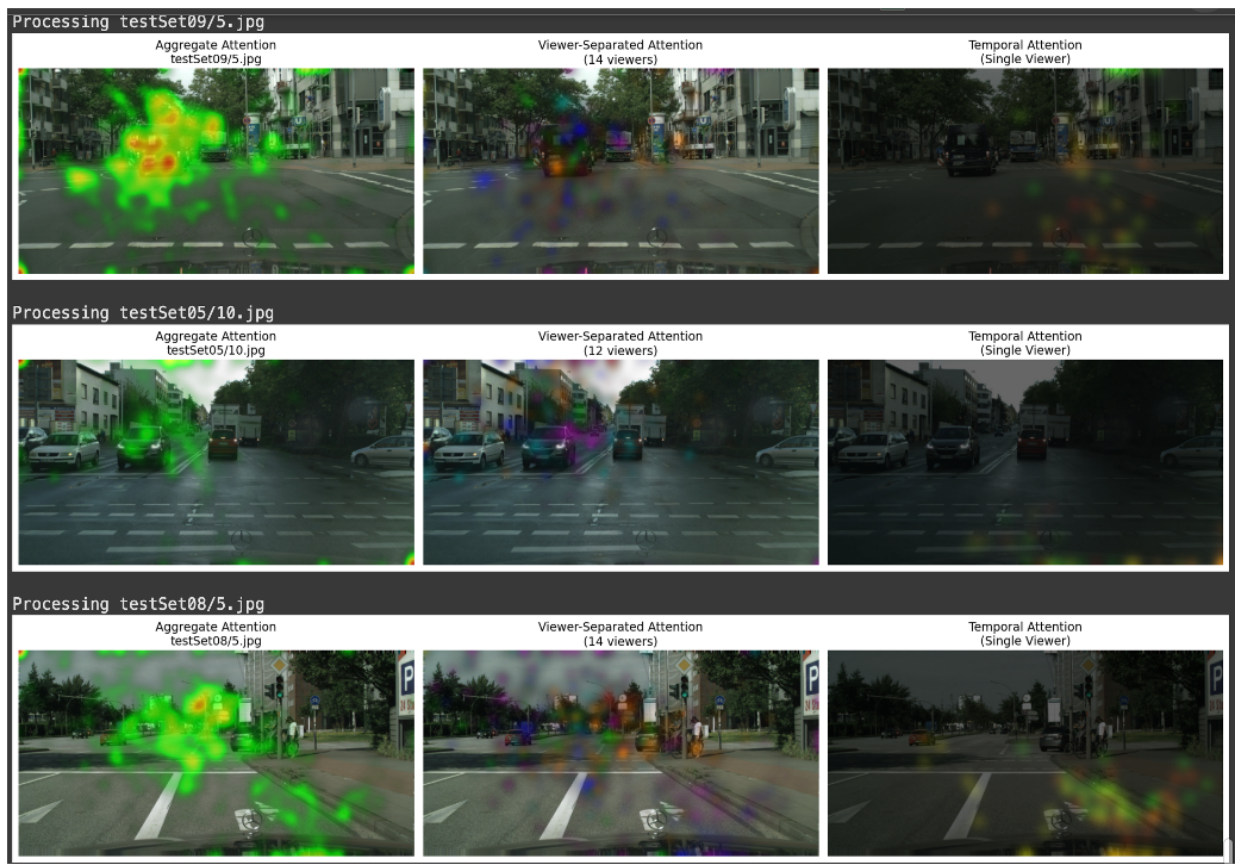

*Figure 1. Saliency maps from the static-image pilot study (Cityscapes dataset). Left column: aggregate attention across all viewers. Center: viewer-separated attention, with each color representing a different participant. Right: temporal attention, where green indicates early fixations and red indicates late fixations. Static images produce coherent spatial attention but cannot capture hazard onset dynamics.*

## 3.2. Source Footage

We collected over 20 hours of front-camera dashcam footage from a 2021 Tesla Model 3 equipped with Hardware 3.0 (HW3), driven through Durham, North Carolina under diverse conditions: rain, snow, dusk, dawn, direct sun, and glare. The HW3 front-facing camera records at $1280 \times 960$ resolution (4:3 aspect ratio). Tesla vehicles record 1-minute segments simultaneously from four cameras; our video ingestion pipeline extracts front-camera footage only, concatenates segments chronologically using FFmpeg, then splits into 15-second clips optimized for participant attention span. The final dataset comprises 388 unique clips synced to AWS S3.

## 3.3. Simulation Platform

We built an interactive driving simulation platform using React/TypeScript with a Node.js backend.[1] Participants view dashcam clips and press the spacebar to mark the start and end of perceived hazardous events. Simultaneously, WebGazer.js (Papoutsaki et al., 2016) captures gaze coordinates via the participant’s webcam at ~10–15 Hz effective sampling rate, depending on browser performance and hardware. Data collection spanned January 2025 through January 2026 across 40 participants, 746 sessions, and 137,663 frame-level gaze samples.

## 3.4. Calibration Protocols

Data collection used two calibration protocols, with the upgrade occurring in October 2025:

- **Pre-calibration** (47,524 gaze samples, 325 videos, 511 sessions): 9-point grid, 5 clicks/point (45 total calibration samples).

[1] Platform available at https://github.com/Onyx-AI-LLC/Human-Alignment-Hazardous-Driving-Detection.

- **Post-calibration** (90,139 gaze samples, 183 videos, 235 sessions): 11-point circle, 40 clicks/point (440 total calibration samples, 10 × more).

This natural split enables a controlled comparison of how calibration quality affects downstream model performance.

## 3.5. ETL Pipeline

All participant data is stored in AWS S3 as JSON files. The ETL pipeline processes data through four stages, each designed to address a specific source of noise:

**Stage 1: Viewport Normalization.** Participants used diverse screen sizes, causing the $1280 \times 960$ video to render at different pixel dimensions with varying letterbox offsets. We compute the exact video display area for each participant's viewport, then transform gaze coordinates through a two-step process: points within the video area are converted to relative positions (0–1 range) within the video content and rescaled to a common $1512 \times 832$ target viewport; points outside the video area (UI elements, letterbox regions) are scaled proportionally. Records with fewer than 25% of gaze points in the video area are flagged as calibration failures.

**Stage 2: Render Delay Removal.** Video playback introduces a variable render delay at session start (buffering, browser initialization). We extract only the final 15 seconds of each session, discarding the initial period where gaze data corresponds to a loading screen rather than driving footage. Spacebar timestamps are re-anchored to this truncated window.

**Stage 3: Demographic Reaction Time Correction.** Hazard annotation timestamps (spacebar presses) include each participant's perceptual-motor reaction time. Following the age-related slowing of choice reaction time documented by Der & Deary (2006) (who found monotonic increases across adulthood in a sample of 7,130 adults), we subtract a per-participant adjustment (clamped to 50–300 ms) from hazard start timestamps. End timestamps at the video boundary are left unadjusted (automatic session termination). This correction shifts hazard windows earlier to better approximate the moment of hazard *perception* rather than hazard *response*. Of 40 participants, 32 provided full demographic profiles; for the remaining 8 (20%), we imputed the cohort mean (age 34, unknown gender, yielding a 70 ms adjustment).

**Stage 4: Frame-Level Feature Extraction.** Session-level records are exploded into individual gaze samples, each enriched with 50+ engineered features: gaze velocity ($v_x$, $v_y$), speed, acceleration ($a_x$, $a_y$), spatial dispersion, fixation/saccade classification, screen region labels (center, peripheral, horizon), distance from center, rolling means, and hazard proximity features (time to nearest spacebar press). The final dataset contains 137,663 frame-level records across 746 sessions.

## 3.6. Feature Engineering

Video frames produce 576-dimensional embeddings via YOLOv8m (Jocher et al., 2023) (Layer 21 SPPF with adaptive average pooling). Nine gaze features—position ($x$, $y$), velocity ($v_x$, $v_y$), speed, and acceleration ($a_x$, $a_y$)—are temporally aligned through a causal module using `merge_asof` with `direction='backward'`, strictly enforcing that only past and present frames inform predictions. All experiments use video-level 80/20 train/test splits to prevent data leakage.

# 4. Experiments

We conduct three experiments with increasing rigor. Each uses 5 random seeds varying the video-level split, with a paired $t$-test to assess statistical significance. The only variable across the baseline and gaze-augmented models is the input features; all hyperparameters are held constant.

## 4.1. Experiment 1a: Random Forest on Pre-Calibration Data

Each sample consists of the current frame embedding plus embeddings from the last 10 gaze samples within a 3-second lookback window. Baseline: $576 \times 11 = 6{,}336$ features. With gaze: $6{,}336 + 9 = 6{,}345$ features.

*Table 1. Experiment 1a: RF on pre-calibration data (9-point, 45 clicks).*

| Seed | Baseline AUC | + Gaze AUC | Δ |
|---|---|---|---|
| 42 | 0.665 | 0.651 | −0.014 |
| 153 | 0.613 | 0.615 | +0.002 |
| 264 | 0.643 | 0.639 | −0.004 |
| 375 | 0.676 | 0.680 | +0.003 |
| 486 | 0.572 | 0.589 | +0.016 |
| **Mean** | **0.634** | **0.635** | **+0.001** |

Paired $t$-test: $p = 0.919$ (**not significant**).

Gaze provides essentially no benefit with the weak calibration protocol. We next investigate the calibration quality.

## 4.2. Calibration Investigation

The pre-calibration protocol used only 45 calibration samples, the minimum WebGazer requires. Papoutsaki et al. (2016) report mean errors of 130–257 px depending on model configuration, with the best model achieving ~ 130 px. Semmelmann & Weigelt (2018) measured webcam accuracy at ~4° visual angle (~172 px) in-lab, degrading further in online settings, compared to 0.5–1° for dedicated infrared eye trackers. Holmqvist et al. (2012) established that gaze data quality must exceed the spatial granularity required by the research task, and Wisiecka et al. (2022) confirmed that calibration quality directly predicts downstream data usability. Post-calibration data (440 samples, 10 × more) shows measurably cleaner signals: lower dispersion, reduced erratic speed, and higher video-area retention (Figure 2).

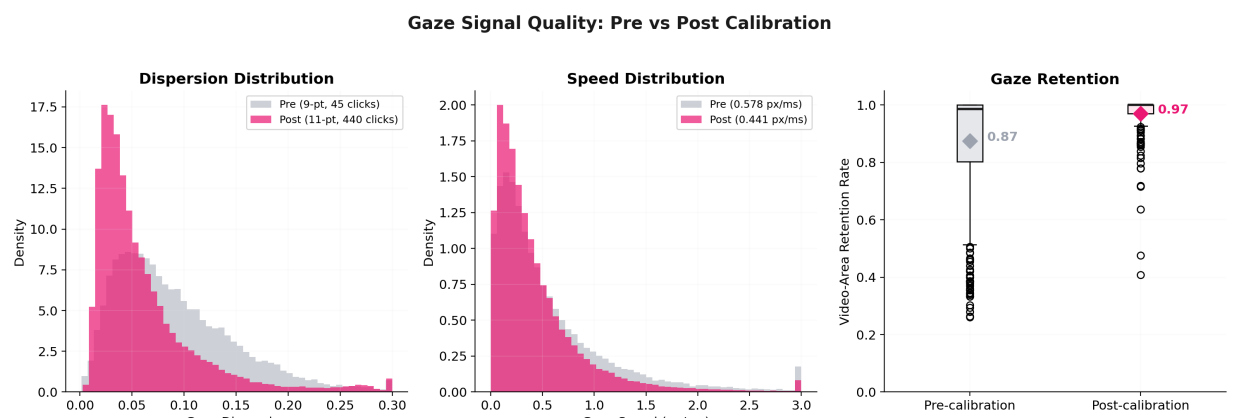

*Figure 2. Gaze signal quality: pre vs. post calibration. Post-calibration shows lower dispersion and higher video-area retention.*

## 4.3. Experiment 1b: Random Forest on Post-Calibration Data

Same setup as 1a, using only post-calibration sessions.

*Table 2. Experiment 1b: RF on post-calibration data (11-point, 440 clicks).*

| Seed | Baseline AUC | + Gaze AUC | Δ |
|---|---|---|---|
| 42 | 0.581 | 0.581 | +0.000 |
| 153 | 0.534 | 0.545 | +0.011 |
| 264 | 0.519 | 0.516 | −0.004 |
| 375 | 0.667 | 0.665 | −0.002 |
| 486 | 0.593 | 0.596 | +0.002 |
| **Mean** | **0.579** | **0.581** | **+0.002** |

Paired $t$-test: $p = 0.578$ (**not significant**).

Better calibration does not rescue the gaze signal; the improvement remains near zero and statistically insignificant. Is the remaining bottleneck model complexity?

## 4.4. Experiment 2: Causal Transformer on Post-Calibration Data

A recent meta-analysis of 74 traffic safety studies found that deep learning architectures outperform traditional ML (including Random Forests) by 7.7 percentage points in F1-score, with temporal models (RNN/LSTM, Transformers) achieving the highest performance (Kotsyubynska et al., 2026). The Transformer architecture (Vaswani et al., 2017) is particularly suited to our task because it learns temporal dependencies directly from sequences through self-attention rather than relying on hand-crafted lag features. To eliminate model complexity as a confound, we train a 2-layer causal Transformer (8.1M parameters) with multi-head self-attention, sinusoidal positional encoding, and padding masks. The model predicts hazards 0.5 s ahead using only past data (strict causality enforced via runtime assertions). Both the baseline (576-dim, 8 heads) and gaze-augmented (585-dim, 9 heads) models use **identical hyperparameters**: LR=5e-5, AdamW, 15 epochs, batch size 32, Xavier initialization (gain=0.5), gradient clipping (max_norm = 1.0), BCEWithLogitsLoss with class-imbalance weighting. Training used CPU for numerical stability.

*Table 3. Experiment 2: Causal Transformer on post-calibration data.*

| Seed | Baseline AUC | + Gaze AUC | Δ |
|---|---|---|---|
| 42 | 0.627 | 0.634 | +0.007 |
| 153 | 0.656 | 0.689 | +0.033 |
| 264 | 0.637 | 0.581 | −0.055 |
| 375 | 0.707 | 0.759 | +0.052 |
| 486 | 0.665 | 0.670 | +0.005 |
| **Mean** | **0.659** | **0.667** | **+0.008** |

Paired $t$-test: $p = 0.667$ (**not significant**).

The Transformer shows the largest mean improvement (+0.008) but with high variance: one seed achieves +0.052 while another goes −0.055. The non-significant $p$-value (0.667) confirms this is noise, not signal.

# 5. Geometric Proof: Object Size vs. Gaze Error

To definitively explain why gaze cannot help at this instrument precision, we measured the rendered size of hazard-relevant objects using YOLOv8m detection (confidence $> 0.3$) on hazard frames, then compared them to WebGazer's error radius.

Across 3,959 detected objects (cars, pedestrians, traffic lights, bicycles, trucks, buses, stop signs) in 481 hazard frames from 50 videos:

- Median object width: **36 px**. WebGazer error: **196 px**. Ratio: **5.5×**.
- **93%** of hazard objects are smaller than the gaze error radius.
- **100%** of pedestrians, traffic lights, and stop signs are smaller.
- One-sample $t$-test (mean width vs. 196 px): $t = -104.4, p < 10^{-6}$.

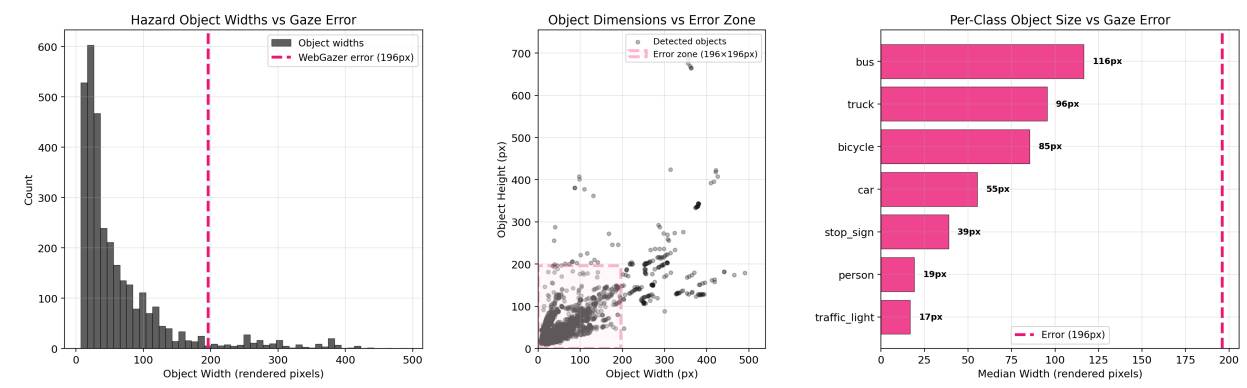

*Figure 3. Hazard object sizes vs. WebGazer error. Left: width distribution with 196 px error line. Center: object dimensions vs. error zone. Right: per-class median width.*

WebGazer's error circle is 5.5× wider than the typical hazard object. Underwood et al. (2003) showed that experienced vs. novice drivers differ primarily in horizontal spread of search fixations, differences that are small relative to webcam error margins, rendering the hazard-gaze signal undetectable at this instrument precision. The instrument cannot distinguish whether a participant is looking *at* a hazard or at the scene *adjacent* to it.

## 6. Summary of Results

*Table 4. All experiments. No combination of calibration quality and model architecture yields statistically significant gaze benefit.*

| Experiment | Data | Mean Δ | $p$ | Sig.? |
|---|---|---|---|---|
| 1a: RF | Pre-cal | +0.001 | 0.919 | No |
| 1b: RF | Post-cal | +0.002 | 0.578 | No |
| 2: Transformer | Post-cal | +0.008 | 0.667 | No |
| Geometric proof | | 93% objects < error ($p < 10^{-6}$) | | |

All three experiments show near-zero mean improvement ($+0.001$, $+0.002$, $+0.008$) with high $p$-values ($0.919$, $0.578$, $0.667$), confirming that gaze provides no detectable benefit at this instrument precision.

## 7. Discussion & Conclusion

We systematically tested whether webcam-based eye tracking improves hazardous driving detection across three axes: calibration quality, model complexity, and statistical rigor. No combination produces a significant improvement. The geometric analysis identifies the root cause: WebGazer's 196 px error exceeds 93% of hazard object sizes, making object-level gaze attribution physically impossible.

**Implications for trustworthy AI.** This work illustrates the importance of *instrument-aware evaluation* in trustworthy AI research. A single-seed Random Forest showed +7.4% AUC improvement, a result that could have been published as a positive finding. Multi-seed validation and root-cause analysis revealed it as noise. The field needs more honest negative results, particularly in safety-critical domains where false positives in evaluation can propagate to deployment decisions.

**The pipeline as a contribution.** While the gaze signal proved insufficient, the data collection infrastructure (simulation platform, 4-stage ETL pipeline, causal alignment module, and multi-seed evaluation framework) is designed for extensibility. Replacing WebGazer with dedicated eye-tracking hardware (Tobii at 0.5–1 ° accuracy) requires only swapping the gaze input layer while retaining the rest of the pipeline. The calibration-era comparison methodology also provides a template for evaluating data quality impact in other sensor modalities.

**Future work.** (1) Dedicated eye-tracking hardware with sub-degree accuracy. (2) Larger post-calibration dataset. (3) Synthetic positive control: inject ground-truth gaze labels to verify the pipeline can detect signal when it exists. (4) Comparison against established driving attention datasets such as DR(eye)VE (Palazzi et al., 2018) and BDD-A (Xia et al., 2018).

## Impact Statement

This paper presents work that evaluates data quality limitations in safety-critical AI systems. Our finding, that webcam eye tracking is insufficient for hazard detection, helps prevent premature deployment of gaze-conditioned driving systems based on insufficient instrument precision. We publish our negative result to save other researchers from the same dead end and to advocate for rigorous multi-seed validation in safety-critical ML.